\documentclass[sigconf]{acmart}
\AtBeginDocument{%
  }

\usepackage{multirow}
\usepackage{makecell}
\usepackage{booktabs}
\usepackage{array} 
\usepackage{graphicx}
\usepackage{rotating}

\usepackage[caption=false,font=footnotesize,labelfont=rm,textfont=rm]{subfig}
\DeclareSubrefFormat{parens}{#1(#2)}
\usepackage[section]{placeins}

\setcopyright{acmlicensed}
\copyrightyear{2026}
\acmYear{2026}
\setcopyright{cc}
\setcctype{by}
\acmConference[WSDM '26]{Proceedings of the Nineteenth ACM International Conference on Web Search and Data Mining}{February 22--26, 2026}{Boise, ID, USA}
\acmBooktitle{Proceedings of the Nineteenth ACM International Conference on Web Search and Data Mining (WSDM '26), February 22--26, 2026, Boise, ID, USA}
\acmPrice{}
\acmDOI{10.1145/3773966.3778007}
\acmISBN{979-8-4007-2292-9/2026/02}
\begin{document}

\title{TimeMAE: Self-Supervised Representations of Time Series with
Decoupled Masked Autoencoders}

\author{Mingyue Cheng}
\affiliation{%
  \institution{State Key Laboratory of Cognitive Intelligence, University of Science and Technology of China}
  \city{Hefei, Anhui Province}
  \country{China}}
\email{mycheng@ustc.edu.cn}

\author{Xiaoyu Tao}
\affiliation{%
\institution{State Key Laboratory of Cognitive Intelligence, University
 of Science and Technology of China}
  \city{Hefei, Anhui Province}
  \country{China}}
\email{txytiny@mail.ustc.edu.cn}

\author{Zhiding Liu}
\affiliation{%
\institution{State Key Laboratory of Cognitive Intelligence, University
 of Science and Technology of China}
  \city{Hefei, Anhui Province}
  \country{China}}
\email{zhiding@mail.ustc.edu.cn}

\author{Qi Liu*}
\affiliation{%
\institution{State Key Laboratory of Cognitive Intelligence, University
 of Science and Technology of China}
  \city{Hefei, Anhui Province}
  \country{China}}
\email{qiliuql@ustc.edu.cn}

\author{Hao Zhang}
\affiliation{%
\institution{State Key Laboratory of Cognitive Intelligence, University
 of Science and Technology of China}
  \city{Hefei, Anhui Province}
  \country{China}}
\email{zh2001@mail.ustc.edu.cn}

\author{Rujiao Zhang}
\affiliation{%
\institution{State Key Laboratory of Cognitive Intelligence, University
 of Science and Technology of China}
  \city{Hefei, Anhui Province}
  \country{China}}
\email{zhangrujiao@mail.ustc.edu.cn}

\author{Enhong Chen}
\affiliation{%
  \institution{State Key Laboratory of Cognitive Intelligence, University of Science and Technology of China}
  \city{Hefei, Anhui Province}
  \country{China}}
\email{cheneh@ustc.edu.cn}

\renewcommand{\shortauthors}{Mingyue Cheng et al.}

\begin{abstract}
  Learning transferable representations from unlabeled time series is crucial for improving performance in data-scarce classification. Existing self-supervised methods often operate at the point level and rely on unidirectional encoding, leading to low semantic density and a mismatch between pre-training and downstream optimization. In this paper, we propose TimeMAE, a self-supervised framework that reformulates masked modeling for time series via semantic unit elevation and decoupled representation learning. Instead of modeling individual time steps, TimeMAE segments time series into non-overlapping sub-series to form semantically enriched units, enabling more informative masked reconstruction while reducing computational cost. To address the representation discrepancy introduced by masking, we design a decoupled masked autoencoder that separately encodes visible and masked regions, avoiding artificial masked tokens in the main encoder. To guide pre-training, we introduce two complementary objectives: masked codeword classification, which discretizes sub-series semantics via a learned tokenizer and masked representation regression, which aligns continuous representations through a momentum-updated target encoder. Extensive experiments on five datasets demonstrate that TimeMAE outperforms competitive baselines, particularly in label-scarce scenarios and transfer learning scenarios.
Our codes are publicly available at \url{https://github.com/Mingyue-Cheng/TimeMAE}.

\end{abstract}





\maketitle

\section{Introduction}
Multivariate time series are prevalent in numerous applications~\cite{choi2024multi, wang2025can}. They consist of temporally ordered observations, with each time step containing measurements across multiple channels~\cite{wang2024tabletime}. Analyzing this type of data in a comprehensive manner substantially improves decision-making across various online applications, including but not limited to anomaly detection and user behavior analysis. Among these tasks, time series classification~\cite{ismail2020inceptiontime,liu2024self} receives significant attention in recent years. A wide range of models is proposed, spanning from early classical methods~\cite{zheng2014time,ye2009time} to recent deep learning-based approaches~\cite{tang2021omni,huang2025msnet}.

In contrast, deep learning methods demonstrate significant advantages due to their ability to scale effectively across diverse time series benchmarks, with substantially less reliance on prior knowledge such as translation invariance~\cite{tonekaboni2021unsupervised}. However, directly applying expressive neural architectures, such as transformer networks~\cite{vaswani2017attention,cheng2023formertime}, to time series classification does not always yield satisfactory results. This less-than-encouraging outcome is partly attributed to the scarcity of annotated data, as transformer architectures heavily rely on large-scale labeled training data~\cite{tonekaboni2021unsupervised}. Annotating time series instances is often time-consuming and labor-intensive, and in certain real-world scenarios, acquiring sufficient annotated time series data is even infeasible.

The challenge of annotating time series data motivates a growing body of work on mining representations from unlabeled datasets. Self-supervised learning~\cite{liu2021self} emerges as a promising paradigm, enabling models to acquire transferable representations without relying on manual labels. In time series analysis, numerous approaches adapt self-supervision principles to capture temporal dynamics and structural patterns~\cite{zhang2023self, cheng2025cross}. Among these, contrastive learning~\cite{hadsell2006dimensionality,choi2024multi} becomes especially popular, aiming to learn embeddings that remain invariant under multi-scale distortions through the combined use of data augmentation and negative sampling strategies~\cite{kong2022understanding}. Nevertheless, such invariance assumptions do not always hold in real-world scenarios, and both augmentation design~\cite{zhang2022self} and negative sampling~\cite{chuang2020debiased} introduce substantial inductive biases. Moreover, these methods predominantly rely on unidirectional encoder architectures, which inherently limit the extraction of contextual information~\cite{devlin2018bert}.

Denoising autoencoders~\cite{bao2021beit, he2022masked} are proposed to effectively address the aforementioned limitations. Their primary idea involves encoding corrupted inputs, using masking-based techniques, into a latent space, followed by recovering the original inputs through an encoder–decoder architecture. A notable example is BERT~\cite{devlin2018bert}, which is built on transformer networks and represents a significant milestone in language representation learning. Inspired by this paradigm, a pioneering effort~\cite{zerveas2021transformer} adopts transformer networks for time series modeling, directly treating raw time series as inputs and recovering the complete signals via point-wise regression objectives. However, this approach often incurs high computational costs, largely due to overlooking the quadratic complexity of self-attention mechanisms~\cite{zhou2021informer,darban2025carla}. As indicated by subsequent studies~\cite{yue2022ts2vec,zhang2022self}, such methods tend to produce limited representations owing to their weak generalization performance.

In our view, the observed insufficient generalization performance primarily arises from a failure to fully account for the distinctive properties of time series. First, in contrast to language data, time series represent natural sequential data characterized by inherent temporal redundancy, such that each time step can often be inferred from its neighboring points. As a result, the pre-trained representation encoder struggles to learn informative representations, since the recovery task becomes overly simplistic. Second, discriminative patterns in time series data frequently manifest as sub-series, commonly referred to as shapelets~\cite{ye2009time,he2023shapewordnet}, implying that individual time steps convey only sparse semantic information. This characteristic leads to a pronounced discrepancy between self-supervised pre-training and downstream task optimization when masking-based strategies are adopted. Specifically, a portion of positions is replaced with masked embeddings during pre-training, whereas such artificial symbols are typically absent in the fine-tuning stage, resulting in a mismatch between the two phases.

To address the aforementioned challenges, we propose a novel masked autoencoder framework, termed TimeMAE, for learning transferable time series representations with transformer networks. Instead of modeling each time step independently, we split each time series into a sequence of non-overlapping sub-series via a window-slicing operation. This design increases the information density of masked semantic units while significantly reducing computational cost and memory consumption due to the shortened sequence length.
After window slicing, we apply masking operations to these redefined semantic units to enable bidirectional encoding of time series representations. We observe that a simple random masking strategy with a relatively high masking ratio of 60\% yields effective representations for time series data. Nevertheless, window slicing and masking inevitably introduce discrepancies between visible and masked positions. To mitigate this issue, we design a decoupled autoencoder architecture, in which the contextual representations of visible (unmasked) and masked regions are extracted separately by two distinct encoder modules.
To guide the pre-training process, we formulate two pretext tasks to facilitate representation recovery. Finally, we conduct extensive empirical experiments on five publicly available datasets to evaluate the effectiveness of TimeMAE. The results clearly demonstrate the effectiveness of the proposed self-supervised paradigm and the decoupled autoencoder architecture.
\section{Related Work}
The goal of time series representation is to adeptly capture and encode temporal patterns within a dataset, thereby facilitating the successful execution of the targeted time series mining task \cite{cheng2025comprehensive,luo2025time}. With regards to classifying time series, extensive research endeavors are devoted to this area, spanning from traditional methodologies~\cite{ye2009time,rakthanmanon2012searching} to the deep learning techniques~\cite{zhang2020tapnet,bagnall2017great}. In our research, our primary focus lies in harnessing the power of the self-supervised learning paradigm to enhance the capacities of existing effective deep learning models, ultimately elevating the overall classification performance~\cite{tao2025values}.

Numerous studies develop self-supervised methods for time series representation learning, in which the current works mainly obey two paradigms: reconstructive~\cite{devlin2018bert,brown2020language} and discriminative~\cite{kiyasseh2021clocs,shi2021self}. The idea of reconstructive methods tries to recover the full input by relying on autoencoders. For example, TimeNet~\cite{malhotra2017timenet} first utilizes an encoder to convert time series into low-dimensional vectors, followed by a decoder that reconstructs the original time series using recurrent neural networks. Another representative reconstructive method is TST~\cite{zerveas2021transformer}, whose main idea is to recover these masked points by the denoising autoencoder based on the transformer architecture. Whereas, the reconstructive optimization is modeled at a point-wise level, leading to very expensive computation consumption and limited generalizability~\cite{eldele2021time,zhang2023self}. 

In contrast, less computation consumption is required in discriminative methods~\cite{eldele2021time}, whose main solution is pulling positive examples together while pushing negative examples away~\cite{lan2022intra}. A common underlying theme among these methods is that they all learn representations with data augmentation strategies~\cite{tian2020makes}, followed by maximizing the similarity of positive examples using a siamese network architecture. The key efforts are devoted to solving the trivial constant solutions~\cite{he2020momentum} via various negative sampling strategies. For instance, in T-Loss~\cite{franceschi2019unsupervised}, time-based negative sampling with triple loss is utilized simultaneously. TNC~\cite{tonekaboni2021unsupervised} leverages the local smoothness of time series signals to treat sub-series neighborhoods as positive examples while regarding non-neighbor regions as negative instances. TS-TCC~\cite{eldele2021time} uses both point-level and instance-level contrasting optimization objectives to align the corresponding representations. 
The distinct characteristic of TS2Vec~\cite{yue2022ts2vec} is hierarchically performing contrasting optimization across multiple scales. Moreover, the assumption of consistency between the time and frequency domains is used in~\cite{zhang2022self}.  However, not only does the data augmentation strategy require many inductive biases, but the invariance assumption also does not always hold.

\section{The Proposed TimeMAE}
In this section,  we begin by providing a formal description of the notations used and outlining the problems under study. Following this, we offer a detailed presentation of the pre-training approach.
\begin{figure*}[h]
	\centering
	\includegraphics[width=0.92\textwidth]{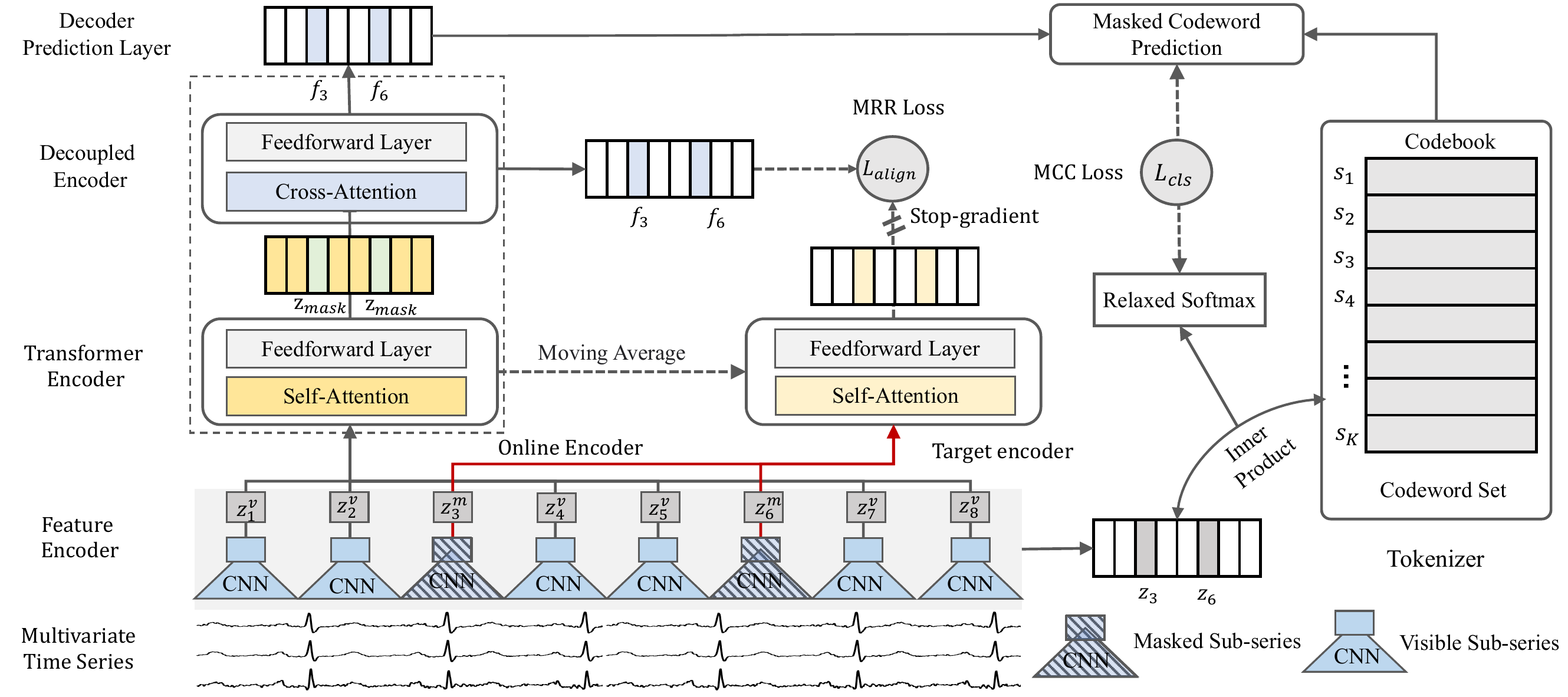} 
	\caption{Illustration of TimeMAE framework: CNN encoders and an online and target transformer jointly process masked and visible sub-series; a decoupled cross-attention decoder then predicts discrete codewords from a learned tokenizer and codebook, trained with relaxed-softmax and momentum-based alignment.
 }
	\label{fig:timemae}
\end{figure*}
\subsection{Problem Definitions}
We begin by introducing the notations and formally defining the problem setting to establish a clear foundation for the subsequent methodology.
$\mathbb{D} = {(X^1, y^1), (X^2, y^2),..., (X^n,y^n)}$ is a dataset containing a collection of pairs $(X^i, y^i)$, in which $n$ denotes the number of examples and $X^i$ denotes the univariate or multivariate time series with its corresponding label denoted by $y^i$. We represent each time series signals as $X= [x_1, x_2, ..., x_T] \in\mathbb{R}^{m\times T}$ where $m$ is the number of channels and $T$ is the number of measurements over time. 
Given a set of unlabeled time series data, self-supervised learning for time series aims to obtain transferable time series representations, which can facilitate the capacity of time series classification tasks. Such representations are expected to capture both local temporal dependencies and global structural patterns, enabling reliable performance across diverse datasets and application scenarios.

\subsection{Feature Encoding Layer}
\subsubsection{Window Slicing Strategies}
To address the challenge of sparse semantic information represented in terms of point-wise units, we propose leveraging the local pattern properties of time series. Instead of employing point-wise inputs, we suggest using sub-series as fundamental modeling elements. As a result, each time series can be processed into a sequence of sub-series units, where each localized sub-series region retains more enriched semantic information and ensures the challenge of the reconstruction task.  To achieve this goal, we employ a \textit{slicing window} operation to segment the raw time series into continuous, non-overlapping sub-series. Formally, for a given time series $X={x_1, x_2, ..., x_T}\in\mathbb{R}^{T\times m}$, a slice is a snippet of the raw time series, defined as $s_{i:j}={x_i,x_{i+1}, ..., x_{i+\sigma}}$. Here, $\sigma$ denotes the size of the slicing window. Assuming the length of the given time series $X$ is $T$, our slicing operation will reduce it to a new length of $\lceil \frac{T}{\sigma} \rceil$. Padding with zeros will be performed to ensure a fixed sequence length. In our implementation, we process the entire time series into a sequence of regular, \textit{non-overlapping} sub-series patches. The primary reason for this choice is that it aims to ensure that the visible regions do not contain information about the reconstructed ones.

Before modeling the sequence dependence of these sub-series units, we encode each element into latent representations using a feature encoder. Here, we employ a 1-D convolution layer to extract local pattern features across channels. It is important to note that different sub-series share the same projection parameters during the feature encoder layer. The projected hidden representations of the input $X$ are denoted by $Z= [z_1, z_2, ..., z_{\lceil \frac{T}{\sigma} \rceil}]\in\mathbb{R}^{\lceil \frac{T}{\sigma} \rceil \times d}$, where $d$ is the dimension of the hidden representation. We use $S$ to denote the new sequence length $\lceil \frac{T}{\sigma} \rceil$, and use $S_v/S_m$ to respectively denote the number of visible/masked positions. 
\subsubsection{Random Masking Strategies}
To acquire comprehensive contextual representations of time series through a bidirectional encoding scheme, we implement masking strategies inspired by the approach outlined in~\cite{devlin2018bert}. Here, let $S_v/S_m$ denote the count of visible (unmasked) /masked positions. This approach allows for the encoding of more extensive contextual information, encompassing preceding and subsequent contexts, for each position simultaneously. To maintain clarity, we use $Z_v$ to represent the embeddings of visible positions, while $Z_m$ corresponds to the embeddings of masked positions. While previous studies, such as block masking~\cite{bao2021beit}, introduce heuristic masking strategies, we adopt random masking strategies to generate corrupted inputs. This means that each sub-series unit has an equal probability of being masked, contributing to the creation of self-supervised signals. This strategy proves effective in ensuring that the representation quality of each input position is substantially improved during the reconstruction optimization. It's important to note that we dynamically mask the time series at random for each pre-training epoch to further enhance the diversity of the training signals. In particular, we encourage a higher masking proportion in TimeMAE. This is pivotal, as the masking ratio significantly impacts the difficulty of the recovery task and, consequently, the encoder's ability to convey more information. A higher masking ratio implies a more challenging recovery task, relying on visible neighboring regions. Consequently, the pre-trained encoder network can encode a more expressive network capacity.  This design choice ultimately encourages the model to learn richer, more generalizable temporal representations that are better suited for downstream time series analysis tasks. 
\subsection{Decoupled Masked Autoencoders}

\subsubsection{Representation of Visible Positions}
In the TimeMAE, we adopt the vanilla transformer encoder architecture denoted by $\mathcal{H}_{\theta}$, consisting of multi-head attention layers and feed-forward network layers, to learn contextual representations of input at visible regions. With such an architecture, the representation of each input unit can obtain the semantic relation of all other positions through the self-attention computation mechanism. It should be noted that the bottleneck of leveraging self-attentive architecture for long sequence input has been largely mitigated by our window slicing operation. Due to the permutation-equivalent self-attention computation, we thus add relative positional embeddings $P\in\mathbb{R}^{ S \times d}$ to each position of sub-series embedding so as to preserve the order of sequence properties. Accordingly, the input embeddings can be reorganized by combining the positional encodings $P$ and the projected representations $Z$, i.e., $Z = Z+P$. Then, we send the input embeddings at visible positions $Z_v$ into the encoder $\mathcal{H}_{\theta}$. There are $L_v$ layers of transformer block in our model, and the output of the last layer $\mathcal{H}_\theta^{L_v} = \{h_1^{L_v}, h_2^{L_v}, ..., h_{S_v}^{L_v}\}$ represents the global contextual representation of $S_v$ visible positions. Unlike previous works, we only feed the representation of visible regions into the encoder network while removing those of masked positions. In this way, no masked embeddings are used to train the encoder network, so that the discrepancy between pre-training and fine-tuning tasks incurred by masked token inputs can be largely alleviated. That is, such strategies eliminate the dilemma of feeding masked tokens into the encoder module in the forward pass.
\subsubsection{Representation of Masked Positions}
To obtain the representations of masked input units, we further replace the self-attention in the vanilla transformer encoder with cross-attention to form the decoupled encoder module, denoted by $\mathcal{F}_{\phi}$, which indicates the decoupling of representation learning between visible and masked inputs. Specifically, we send these representations at visible and masked positions into the decoupled encoder so as to represent the input units at masked positions. Note that the embedding at masked positions is replaced with a newly initialized vector $z_{mask}\in\mathbb{R}^d$ while keeping the corresponding positional embedding unchanged. For clarity, embeddings of the reinitialized masked regions are indicated by $\tilde{Z}$. During representation learning in the decoupled encoder, we treat the representations of masked positions as query embeddings, i.e., $\tilde{Z}$, while regarding the transformed embeddings of visible positions $\mathcal{H}_{\theta}^{L_v}$ as input to help form keys and values. Formally, the model parameters of masked queries $W^{Q_{m}}\in\mathbb{R}^{d\times d}$ are the same for all masked queries. The projection parameters of visible keys and values, i.e., $W^{K_{v}}, W^{V_{v}}\in\mathbb{R}^{d\times d}$ are also shared with visible inputs. The $L_m$-th layer output of the decoupled encoder $\mathcal{F}_{\phi}^{L_m} = \{f_1^{L_m}, f_2^{L_m}, ..., f_{S_m}^{L_m} \}$ represents the transformed contextual representations of $S_m$ masked positions. Note that the decoupled module solely makes predictions of embeddings of masked positions while maintaining the embeddings of visible ones not updated. The main reason is that we hope such an operation can help alleviate the discrepancy issue in the backward pass. By this segmentation operation, the representation role of visible input is only taken responsibly by the previous encoder $\mathcal{H}_\theta$.  Meanwhile, the decouple encoder mainly focuses on the representation of masked positions. Furthermore, another advantage of the decoupled module $\mathcal{F}_{\phi}$ is to prevent the decoder prediction layer from making the representation learning of visible positions, so that the encoder module $\mathcal{H}_\theta$ can carry more meaningful information.  

\subsection{Self-supervised Optimization}

\subsubsection{Masked Codeword Classification}
With the window slicing operation, each time series is reformulated into a sequence of sub-series, which could exhibit more enriched semantics. Taking inspiration from product quantization~\cite{jegou2010product},  a natural idea is whether we can represent such reformulated series in a novel discrete view, i.e., assign each local sub-series with its own ``codeword''. Then, these assigned codewords serve as surrogate supervision signals for missing parts. Hence, we decide to develop a \textit{tokenizer} module, which can convert the continuous embeddings of masked positions into discrete codewords in an \textit{end-to-end} manner.  We name such type of reconstruction task masked codeword classification (MCC). 

To be precise, we assume that the tokenizer is composed of a codebook matrix $C=\{c_1, c_2, ..., c_K\}\in \mathbb{R}^{K\times d}$, consisting of $K$ latent vectors. Here, $c_k\in\mathbb{R}^d$ denotes the $k$-th codeword in the codebook. In addition, the indices of the codebook naturally form the supervision vocabulary $V = \{v_1, v_2, ..., v_K\}$. The key idea is to assign the nearest codeword to each sub-series through similarity computation. Concretely, the codeword of each masked sub-series representation $z_i \in\mathbb{R}^d$ is approximately encoded as follows,
\begin{equation}
	\label{equ:max}
	z^m_i \approx v_{k_i}, s.t. k_i = \arg \max_{j} sim(z_i, c_j), j\in[K],
\end{equation}\noindent 
where $sim (z_i, c_j)$ is a measurement function to evaluate the similarity between sub-series representation and candidate codeword vectors. Instead of using cosine similarity, we use \textit{inner product} to estimate the relevance score in the tokenizer, since the gradient explosion incurred the reciprocal of the norm can be easily prevented. In this way, each local sub-series can be assigned its own discrete codeword, representing the inherent temporal patterns. After assigning the codeword to each input unit, we pass the embedding matrix to the decoder layer to obtain the codeword prediction distribution $p(s_{k_i}|f_i^{L_m})$ so as to perform the MCC optimization. Although heavy decoders might result in greater generation capability, their great capability can only serve for prediction in pretext tasks and cannot benefit the target tasks during the fine-tuning process. Hence, we abandon the heavy decoder head design and only use the extremely lightweight prediction head --- inner product as a measurement tool to obtain the similarity distribution between contextual representation and candidate codeword vectors.

For simplicity, we adopt cross-entropy loss to form the class token recovery optimization, denoted as $\mathcal{L}_{cls}$:
\begin{equation}
	\label{equ:ce}
	\mathcal{L}_{cls}(q, p) = - q(s_{k_i}|z_i) \log p(s_{k_i}|f_i^{L_m})
\end{equation}\noindent where $q(s_{k_i}|z_i)\in[0,1]^K$ denotes one-hot encoding obtained by the maximum selection of in codeword assignment. The optimization goal in Equation~\ref{equ:ce} is equal to maximizing the log-likelihood of the correct codewords given the masked input embeddings. However, it is reported that this codeword maximum selection operator easily leads to two aspects of issues: (1) it is easy to lead to the collapse results~\cite{cheng2025cross}, i.e., only very few ratios of codewords are selected; (2) it makes it non-differentiable for the optimization loss in the above equation so that the back-propagation algorithm cannot be applied to computing gradients.

To solve these issues, we hence relax the maximum by tempered softmax combined with a prior distribution. Formally, the probabilities of selecting $k$-th codeword  can be described as follows
\begin{equation}
	q(v_k|z_i) \approx \tilde{q}(v_k|z_i) = \dfrac{\exp((sim(z_i, c_k)+n_k)/\tau)}{\sum_{k'\in K} \exp((sim(z_i, c_{k'})+n_{k'})/\tau)} ,
\end{equation} \noindent in which $\tau$ is a temperature factor, controlling the degree of approximation. For example, the approximation becomes exact if $\tau\rightarrow 0$, whereas $\tilde{q}$ is close to the uniform distribution when $\tau$ is too large. In addition, $n = -\log(-\log(a))$ and $a$ is uniform samples from $\mathcal{U}(0, 1)$. In our implementation, we adopt Gumbel prior distribution to implement $\mathcal{U}$ by following~\cite{jang2016categorical}. With such a designed relaxed softmax, the maximum selection is replaced by approximately sampling from the probability distribution so that the collapse issue can be largely alleviated. 

To solve the non-differentiable issue, by following the trick of Straight-Through Estimator (STE) in~\cite{bengio2013estimating}, we mimic the gradient updating by rewriting $b_j$ as follows,
\begin{equation}
	\hat{q}(v_k|z_i) = \tilde{q}(v_k|z_i) + sg (q(s_k|z_i) - \tilde{q}(s_k|z_i)),
\end{equation}\noindent where $sg$ denotes the stop gradient operator, i.e., zero partial derivatives, which can effectively constrain its operation to be a non-updated constant. In the forward pass, the stop gradient does not work, i.e., $\hat{q}(v_k|z_i) = q(v_k|z_i) = $$\rm one\_hot$$ (\arg\max_j sim(z_i, c_j))$. During the forward pass, the index of the nearest embedding is assigned as the current embedding vector's discrete codeword. During the backward pass, the stop-gradient operation takes effects, i.e., $\nabla_{\hat{q}(v_k|z_i)} = \nabla_{\tilde{q}(v_k|z_i)}$. Accordingly, the gradient $\nabla \mathcal{L}_{cls}$ is passed unaltered to the embedding space of the codebook matrix, which means that the non-differentiable issue is solved. 

\subsubsection{Masked Representations Regression}
To generate the target representations, a novel \textit{target encoder} module is further adopted. Let $\mathcal{H}_\xi$ denote the target encoder, which preserves the same hyper-parameter setting as the encoder of $\mathcal{H}_\theta$ but parameterized with $\xi$. To facilitate the understanding, we name the combination of the vanilla transformer encoder and the decoupled encoder modules as \textit{online encoder}, in which aligned representations are generated. Relying on such a siamese network architecture, different views of the masked sub-series representations can be produced by the target encoder and online encoder, respectively. Accordingly, aligning the corresponding representation of two different views, i.e., target representations $\mathcal{H}_{\xi}^{L_v}(Z_m)$ and predicted representations $\mathcal{F}_{\phi}^{L_m}(\tilde{Z})$, naturally form the goal of MRR task. Given continuous representation as regression signals, we adopt the mean squared error (MSE) loss to form the specific optimization objectives. Subsequently, we define the exact regression optimization loss as follows, 
\begin{equation}
	\label{equ:align}
	\mathcal{L}_{align} = \parallel\mathcal{H}_{\xi}^{L_v}(Z_m)- \mathcal{F}_{\phi}^{L_m}(\tilde{Z})\parallel_2^2.
\end{equation}\indent Theoretically, the gradient of loss in Equation~\ref{equ:align} can be employed to optimize both the online encoder and target encoder. Since the negative example is omitted in the siamese network architecture, however, it is easy to incur collapse solution~\cite{cheng2025cross}. To prevent the model collapse results, we can obey the rule of updating two networks differently, whose effectiveness has been proved to be effective in previous works~\cite{chen2020simple}. Following this idea, we perform a stochastic optimization step to only minimize the parameters of the online encoder while updating the target encoder in a momentum moving-average manner. 

Particularly, we describe the update of the target encoder as 
\(\xi \leftarrow \eta \cdot \xi + (1-\eta)\cdot\theta,\;
\theta \leftarrow \theta - \gamma\nabla_\theta \mathcal{L}_{align},\;
\phi \leftarrow \phi - \gamma\nabla_\phi \mathcal{L}_{align},\)
in which \(\eta \in [0,1]\) denotes the momentum coefficient for the moving-average update.
With a large value of $\eta$, the target encoder slowly approximates the encoder $\mathcal{H}_\theta$.  In the TimeMAE, we find $\eta = 0.99$ performs effectively. Further, $\nabla_{\cdot}$ is the gradient and  $\gamma$ denotes the learning rate for stochastic optimization. The direction of updating target encoder $\mathcal{H}_\xi$ completely differs from that of updating the $\mathcal{H}_\theta$ and $\mathcal{F}_\phi$. Finally, $\mathcal{H}_{\xi}$ converges to equilibrium by the slow-moving average. To be more specific, at each training step, the optimization direction of the online encoder and target encoder are decided by gradient produced by $\mathcal{L}_{align}$ and $\xi-\theta$. By doing so, the problem of model collapse can be effectively solved with such different updating manners. 
In the pre-training stage, the TimeMAE is trained in a multi-task manner by combining the MCC and MRR tasks together. Precisely, the overall self-supervised optimization objectives of the pre-training model can be written as, 
\begin{equation}
	\label{equ:pre-training}
	\mathcal{L} = \alpha \mathcal{L}_{cls} + \beta \mathcal{L}_{align}, 
\end{equation}\noindent where $\alpha$ and $\beta$ are the tuned hyper-parameters, jointly controlling the weight of the two losses.

\section{Experiments}
\label{sec:experimental_setup}
\begin{table}[h]
	\centering
	\caption{Statistics of the five datasets used in experiments.}
	\vspace{-0.1in}
	\begin{tabular}{ccccc}
		\toprule
		Dataset & \# Example & Length & \# Channel & \# Class \\
		\midrule
		HAR   & 11,770 & 128   & 9     & 6 \\
		PS & 6,668  & 217   & 11    & 39 \\
		AD & 8,800  & 93    & 13    & 10 \\
		Uwave & 4,478  & 315   & 3     & 8 \\
		Epilepsy & 11,500 & 178   & 1     & 2 \\
		\bottomrule
	\end{tabular}%
	\vspace{-0.1in}
	\label{tab:datasets}%
\end{table}
\subsection{Experimental Settings}
\subsubsection{Datasets}
We conduct experiments on five publicly available datasets, including human activity recognition (HAR), PhonemeSpectra (PS), ArabicDigits (AD), Uwave, and Epilepsy. These datasets, which are provided by~\cite{andrzejak2001indications,anguita2013public,eldele2021time,liu2021gated}, cover a broad range of variations: different numbers of channels, varying time series lengths, different sampling rates, diverse scenarios, and diverse types of time series signals. It should be noted that we do not perform any additional processing on these datasets, as the original training and testing sets are already well prepared. Furthermore, Table~\ref{tab:datasets} summarizes the key statistics of each dataset.
including the number of instances (\# Instance), the sequence length (\# Length), the number of channels (\# Channel), and the number of classes (\# Class). 
\begin{table*}[h]
	\centering
	\caption{Comparison of the TimeMAE against competitive baselines in terms of both FineLast and FineAll across five datasets.}
	\setlength{\tabcolsep}{0.8mm}{
		\begin{tabular}{c|ccccc|ccccc}
			\toprule
			Metrics & \multicolumn{5}{c|}{FineLast \& Accuracy}       & \multicolumn{5}{c}{FineAll \&Accuracy} \\
			Datasets & HAR    & PS    & AD    & Uwave & Epilepsy & HAR    & PS    & AD    & Uwave & Epilepsy \\
			\midrule
			FineZero & 68.02±0.84 & 6.92±0.34 & 80.27±3.18 & 88.35±0.27 & 92.26±0.13 & 89.73±0.42 & 8.98±0.54 & 98.68±0.09 & 97.38±0.25 & 97.14±0.13 \\
			FineZero+ & 66.53±0.70 & 7.98±0.27 & 69.45±3.95 & 87.56±0.28 & 95.05±0.31 & 92.26±0.47 & 17.03±0.52 & 98.53±0.18 & 97.35±0.23 & 98.92±0.07 \\
			TST   & 87.39±0.27 & 8.93±0.58 & \textbf{96.03±0.44} & 95.17±0.51 & 96.19±0.09 & 94.35±0.11 & 8.82±0.30 & 99.06±0.39 & 97.98±0.12 & 96.91±0.11 \\
			TNC   & 87.82±0.18 & 10.05±0.52 & 92.38±1.27 & 91.36±0.60 & 97.17±0.09 & 92.02±0.75 & 10.58±0.25 & 98.24±0.47 & 97.06±0.20 & 97.43±0.05 \\
			TS-TCC & 77.63±0.20 & 8.88±0.39 & 88.42±1.52 & 91.76±0.43 & 95.21±0.15 & 89.22±0.19 & 8.47±0.68 & 98.71±0.09 & 97.32±0.28 & 96.83±0.05 \\
			TS2Vec & 78.16±0.80 & 10.82±0.38 & 94.65±0.30 & 92.76±0.34 & 97.13±0.15 & 86.98±2.17 & 10.07±0.66 & 99.11±0.23 & 96.69±0.21 & 96.99±0.11 \\
			SimMTM & 74.86±0.50 & 7.81±0.40 & 91.18±0.20 & 73.44±0.50 & 65.94±0.60 & 93.96±0.30 & \textbf{23.83±0.50} & 99.16±0.05 & 91.25±0.40 & 97.83±0.20 \\
			TimeMAE & \textbf{91.31±0.10} & \textbf{19.25±0.22} & 95.76±0.51 & \textbf{95.88±0.28} & \textbf{97.88±0.20} & \textbf{95.11±0.18} & 20.30±0.27 & \textbf{99.20±0.03} & \textbf{98.37±0.12} & \textbf{99.35±0.19} \\
			\midrule
			& \multicolumn{5}{c|}{FineLast \& F1 Score}       & \multicolumn{5}{c}{FineAll \& F1 Score} \\
			\midrule
			FineZero & 66.39±1.04 & 6.18±0.31 & 80.10±3.13 & 88.27±0.28 & 87.99±0.20 & 89.72±0.39 & 8.47±0.50 & 98.68±0.09 & 97.38±0.25 & 95.54±0.18 \\
			FineZero+ & 64.65±0.83 & 7.44±0.30 & 68.95±4.24 & 87.46±0.28 & 92.12±0.49 & 92.27±0.46 & 15.33±0.43 & 98.53±0.19 & 97.35±0.24 & 98.33±0.10 \\
			TST   & 87.18±0.29 & 7.71±0.78 & 96.02±0.44 & 95.16±0.51 & 93.96±0.20 & 94.33±0.07 & 7.79±0.91 & 99.06±0.39 & 97.98±0.13 & 95.18±0.16 \\
			TNC   & 87.63±0.19 & 8.72±0.22 & 92.35±1.29 & 91.34±0.60 & 95.64±0.15 & 91.89±0.79 & 9.26±0.52 & 98.24±0.47 & 97.05±0.20 & 96.04±0.07 \\
			TS-TCC & 77.09±0.14 & 7.54±0.35 & 88.41±1.55 & 91.71±0.48 & 92.34±0.22 & 89.04±0.15 & 8.00±0.54 & 98.71±0.09 & 97.34±0.28 & 95.09±0.07 \\
			TS2Vec & 77.43±0.91 & 10.09±0.26 & 94.64±0.31 & 92.71±0.33 & 95.58±0.25 & 86.69±2.38 & 8.63±1.06 & 99.10±0.23 & 96.70±0.21 & 95.26±0.22 \\
			SimMTM & 73.65±0.45 & 4.26±0.35 & 91.18±0.20 & 60.73±0.55 & 59.57±0.65 & 94.11±0.30 & \textbf{23.52±0.50} & 99.16±0.05 & 91.23±0.40 & 97.80±0.20\\
			TimeMAE & \textbf{91.25±0.06} & \textbf{18.18±0.27} & \textbf{95.75±0.51} & \textbf{95.87±0.29} & \textbf{96.66±0.35} & \textbf{95.10±0.17} & 18.83±0.72& \textbf{99.20±0.03} & \textbf{98.39±0.12} & \textbf{98.85±0.24} \\
			\bottomrule
	\end{tabular}}
	\label{tab:cls}%
\end{table*}%
\subsubsection{Compared Baselines}
To demonstrate the effectiveness of TimeMAE, we compare it with several popular self-supervised time series representation methods. Specifically, FineZero trains vanilla transformer encoders from scratch without self-supervised pre-training. TST~\cite{zerveas2021transformer} processes time series point-wise and formulates raw series regression as a self-supervised task.  TNC~\cite{tonekaboni2021unsupervised} uses a debiased contrastive objective to distinguish neighborhood from non-neighborhood signal distributions. TS-TCC~\cite{eldele2021time} jointly optimizes point-level and instance-level contrastive objectives. 
TS2Vec~\cite{yue2022ts2vec} performs hierarchical contrastive optimization over augmented views at multiple scales. SimMTM~\cite{dong2023simmtm} combines masked modeling with manifold learning, recovering masked points by a weighted aggregation of multiple neighbors to preserve local structure.
In addition to the comparison baselines listed above, for a fair comparison, we replace the point-wise input processing in the training encoder with newly designed window slicing strategies, denoted by \textbf{FineZero+}. 
Because our goal is to compare the performance of self-supervised learning models independent of encoder settings, we follow~\cite{tonekaboni2021unsupervised} and use the same encoder as in~\cite{devlin2018bert} for all methods.

\subsubsection{Pre-training and Fine-tuning Setup}
In our experiments, we demonstrate the effectiveness of pre-trained model parameters with two mainstream evaluation manners: \textbf{linear evaluation} and \textbf{fine-tuning evaluation}. The former manner means the parameters of the pre-trained encoder are frozen, whereas only the newly initialized classifier layer is tuned according to the labels of target tasks. The latter manner denotes that the parameters of pre-trained encoders and the newly initialized classifier layer are tuned without any frozen operation. These two evaluation manners are denoted by \textbf{FineLast} and \textbf{FineAll}, respectively. Relatively speaking, the measurement of FineLast manner could further reflect the quality of pre-trained models.  For the time series classification task, we require the entire instance-level representation. As such, we use mean pooling over all temporal points to denote the whole time series, followed by a cross-entropy loss to guide the optimization of models in the training stage of downstream tasks. To better evaluate the imbalanced datasets, the metrics of accuracy and macro-averaged F1 Score are employed as the evaluation metrics of time series classification tasks. The best results are highlighted in \textbf{boldface}. In default settings, all results of Accuracy and F1 Score in this table are denoted in the percentage (\%). \textbf{Reported experimental results are mean and standard deviation values across three independent runs on the same data split}. 

\subsubsection{Implement Details}
In the pre-training stage, we consistently regard the embedding size as $64$ in all models. Adam optimizer is employed as the default optimizer for all compared methods. The learning rate is set to $0.001$ without any additional tricky settings. The batch size is set to $64$ for all models. By default, the layer depth of the utilized transformer encoder architecture is set to $8$, in which the number of attention heads is $4$, a two-layer feed-forward network is adopted, and the dropout rate is $0.2$. For several competitive baselines, we strictly follow the corresponding hyper-parameter settings and data augmentation strategies suggested by the original work~\cite{zerveas2021transformer,tonekaboni2021unsupervised,eldele2021time,yue2022ts2vec}. In addition to the used transformer encoder, we also use a six-layer decoupled encoder to extract contextual representations of masked positions. For each dataset, the used slicing window size $\delta$ is searched from $\{4, 8, 12, 16\}$.  In the default setting, we adopt the masking ratio of $60\%$ for our TimeMAE model. Within the MCC task, the vocabulary size of the codebook in our designed tokenizer is searched from  $\{64, 96, 128, 192, 256, 512\}$ for the specific dataset. As for the MRR task, the $\tau$ is set to $0.99$. To obtain the optimal trade-off between two pretext tasks, we fix the $\alpha$ as $1$ while searching the suitable values of $\beta$ ranging from $\{1, 2, ..., 10\}$. Note that we preserve the same hyper-parameter settings as the pre-trained model while consistently involving the overlapping hyper-parameters in the fine-tuning stage for fairness.

\begin{table*}[htbp]
	\centering
	\caption{Transfer learning performance comparison between the TimeMAE and several baselines under the FineAll evaluation setting. The best results in each column are highlighted in boldface.}
	\setlength{\tabcolsep}{1.0mm}{
	\begin{tabular}{c|cccc|cccc}
		\toprule
		\multirow{2}[2]{*}{Models} & \multicolumn{4}{c|}{Accuracy} & \multicolumn{4}{c}{F1 Score} \\
		& PS & AD & Uwave & Epilepsy & PS & AD & Uwave & Epilepsy \\
		\midrule
		FineZero & 8.98±0.54 & 98.68±0.09 & 97.38±0.25 & 97.14±0.13 & 8.47±0.50 & 98.68±0.09 & 97.38±0.25 & 95.54±0.18 \\
		FineZero+ & 7.98±0.27 & 69.45±3.95 & 87.56±0.28 & 95.05±0.31 & 17.03±0.52 & 98.53±0.18 & 97.35±0.23 & 98.92±0.07 \\
		TST   & 9.43±0.31 & 98.71±0.21 & 97.24±0.48 & 97.18±0.08 & 7.77±0.34 & 98.71±0.22 & 97.26±0.48 & 95.61±0.08 \\
		TNC   & 9.43±0.75 & 98.56±0.05 & 96.98±0.56 & 97.07±0.18 & 8.28±0.79 & 98.56±0.05 & 96.99±0.55 & 95.48±0.28 \\
		TS-TCC & 9.46±0.28 & 98.50±0.08 & 97.48±0.08 & 97.11±0.13 & 8.20±0.37 & 98.50±0.08 & 97.48±0.07 & 95.54±0.16 \\
		TS2Vec & 6.52±0.10 & 98.62±0.12 & 96.85±0.14 & 96.73±0.22 & 4.30±0.10 & 98.62±0.12 & 96.86±0.12 & 94.82±0.25 \\
        SimMTM & \textbf{22.01±0.40}& 97.83±0.20 & 92.50±0.35 & 96.16±0.05 & \textbf{21.93±0.38} & 97.87±0.22 & 92.54±0.33 & 96.86±0.04 \\
		TimeMAE & 18.53±0.69 & \textbf{98.76±0.07} & \textbf{97.72±0.08} & \textbf{99.08±0.12} & 17.21±0.43 & \textbf{98.76±0.07} & \textbf{97.73±0.08} & \textbf{98.76±0.16} \\
		\bottomrule
	\end{tabular}}
	\label{tab:tl}%
\end{table*}

\subsection{Experimental Results}
\subsubsection{One-to-One Pre-training Evaluation}
In this section, we demonstrate the effectiveness of our approach against the competitive baselines in terms of the one-to-one pre-training paradigm. Specifically, we first perform self-supervised pre-training without using annotated labels, followed by a retraining stage tuned with labeled datasets for classification tasks. Finally, the model's performance is evaluated in the testing set. 

Table~\ref{tab:cls} shows the classification results. We observe that the classification performance of the compared methods evaluated by FineAll manner can be more promising than that evaluated by FineLast manner.
This is reasonable because the FineAll evaluation fully forces the whole model parameter to adapt to target tasks. In contrast, only the last classifier is tuned according to the target label in the evaluation manner of FineLast. In both metrics, our TimeMAE achieves substantial performance improvements compared to other baselines in most situations. In particular, the capacity of our TimeMAE in terms of linear evaluation could outperform supervised learning version baselines on the HAR and PS datasets, which fully demonstrates the powerful representation learning capacity of our TimeMAE model. We attribute such success to the simple philosophy of recovering missing sub-series patches based on the visible inputs without relying on too much inductive bias, like the data augmentation strategies in contrastive-based baselines.  Among these baselines, it seems that there exist no apparent winners for all datasets. It is worth noting that masking-based autoencoders like baseline---TST cannot achieve promising improvements in contrast to our TimeMAE model. We hypothesize that the performance degradation is mainly led by the sparse basic semantic unit in processing the time series in a point-wise style. Such arguments can also be evidenced by the comparison of results between FineZero and FineZero+, further confirming the effectiveness of our design.

\begin{figure*}[t]
	\centering
	\includegraphics[width=1.0\textwidth]{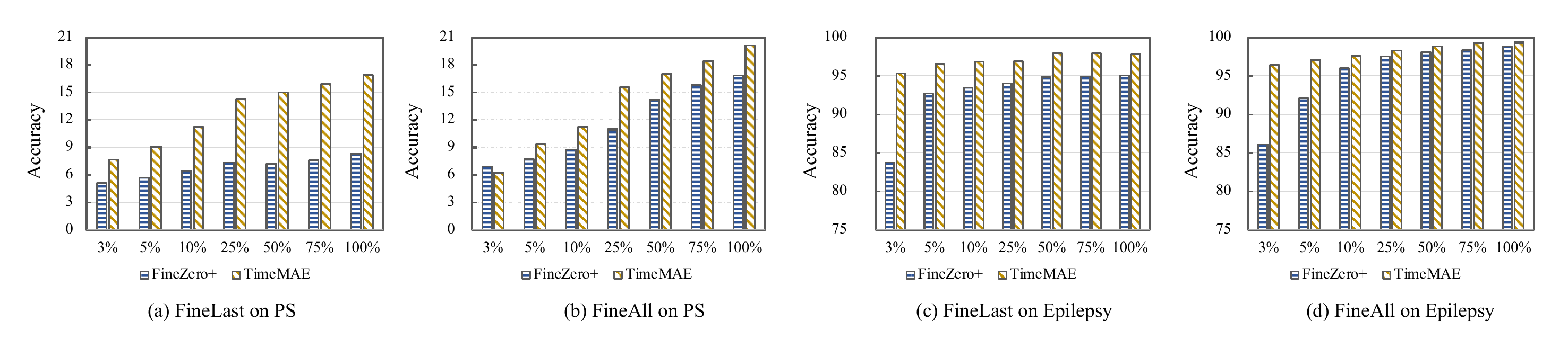} 
	\caption{Comparison between FineZero+ v.s. our TimeMAE for different proportions of training set over two selected datasets, including PS and Epilepsy, in which both experimental results of FineLast and FineAll are reported. }
	\label{fig:ratio_training}
\end{figure*}

\begin{table*}[htbp]
	\centering
	\caption{Performance analysis of the TimeMAE model with respect to varying sizes of encoder network. }
	\vspace{-0.08in}
	\setlength{\tabcolsep}{2.8mm}{
	\begin{tabular}{c|c|c|cccccc}
		\toprule
		Evaluation  & Datasets & Model Size &  
		\makecell[c]{l = 8 \\ d = 64 \\ n = 100} &
		\makecell[c]{l = 8 \\ d = 64 \\ n = 200} &
		\makecell[c]{l = 8 \\ d = 128 \\ n = 200} &
		\makecell[c]{l = 16 \\ d = 64 \\ n = 200} &
		\makecell[c]{l = 16 \\ d = 128 \\ n = 100} &
		\makecell[c]{l = 16 \\ d = 128 \\ n = 200} \\
		\midrule
		\multirow{6}[6]{*}{FineLast} & \multirow{2}[2]{*}{HAR} & Accuracy & 91.31±0.10 & 91.99±0.12 & 92.63±0.64 & 91.97±1.81 & 91.40±0.44 & \textbf{92.82±0.31} \\
		&       & F1 Score & 91.25±0.06 & 92.06±0.11 & 92.63±0.66 & 91.97±1.83 & 91.41±0.42 & \textbf{92.42±0.30} \\
		\cmidrule{2-9}          
		& \multirow{2}[2]{*}{PS} & Accuracy & 16.91±1.26 & 20.90±0.55 & 24.27±0.33 & 22.61±0.13 & 21.84±0.69 & \textbf{25.34±0.98} \\
		&       & F1 Score & 15.72±1.19 & 19.81±0.54 & 23.88±0.36 & 22.00±0.17 & 21.39±0.62 & \textbf{25.02±0.84} \\
		\cmidrule{2-9}          
		& \multirow{2}[2]{*}{Epilepsy} & Accuracy & 97.88±0.20 & 98.56±0.16 & 99.04±0.14 & 98.91±0.18 & 98.88±0.08 & \textbf{99.33±0.06} \\
		&       & F1 Score & 96.66±0.35 & 97.76±0.25 & 98.51±0.22 & 98.31±0.29 & 98.26±0.13 & \textbf{98.87±0.09} \\
		\midrule
		\multirow{6}[6]{*}{FineAll} & \multirow{2}[2]{*}{HAR} & Accuracy & 95.11±0.18 & 94.98±0.75 & 95.19±0.16 & 95.43±0.52 & 95.08±0.08 & \textbf{95.51±0.65} \\
		&       & F1 Score & 95.10±0.17 & 94.95±0.74 & 95.15±0.20 & 95.42±0.50 & 95.03±0.10 & \textbf{95.02±0.69} \\
		\cmidrule{2-9}          
		& \multirow{2}[2]{*}{PS} & Accuracy & 20.15±0.50 & 22.97±0.45 & 23.28±0.21 & 23.43±0.27 & 22.35±0.15 & \textbf{24.81±1.49} \\
		&       & F1 Score & 18.86±0.52 & 21.97±0.48 & 22.15±0.95 & 22.37±0.78 & 20.76±0.40 & \textbf{23.97±1.52} \\
		\cmidrule{2-9}          
		& \multirow{2}[2]{*}{Epilepsy} & Accuracy & 99.35±0.19 & 99.46±0.11 & 99.41±0.19 & \textbf{99.57±0.01} & 99.41±0.08 & 99.52±0.03 \\
		&       & F1 Score & 98.85±0.24 & 99.16±0.17 & 99.07±0.29 & \textbf{99.32±0.01} & 99.07±0.13 & 99.28±0.05 \\
		\bottomrule
	\end{tabular}}
	\label{tab:scable_analysis}%
	\vspace{-0.08in}
\end{table*}

\subsubsection{One-to-Many Pre-training Evaluation}
Recent advancements in self-supervised methods~\cite{brown2020language} show great potential for learning universal representations. Hence, to verify the generic capacity of the pre-trained model,  we seek to investigate the transferability~\cite{Long2016DeepLO} of the pre-trained model by following a one-to-many evaluation. To be more specific, self-supervised pre-training is done by using only one dataset, followed by fine-tuning multiple different target datasets. Out of five datasets, the HAR dataset has the largest number of examples. For that reason, we perform self-supervised pre-training on the HAR dataset and separately fine-tune the well-trained model on the remaining four datasets, including PS, AD, Uwave, and Epilepsy.  
As shown in Table~\ref{tab:tl}, we report classification performance in terms of FineLast and FineAll. We observe that the overall fine-tuning performance of TimeMAE is slightly worse than in the one-to-one evaluation setting. This is reasonable because there are fewer shared characteristics between HAR and the other datasets, so transfer learning is less effective than fine-tuning within the same dataset. Among these methods, TimeMAE surpasses all competitive baselines and achieves strong performance in transfer learning. Meanwhile, some pre-trained representation models also benefit the target task by transferring learned parameters. We hypothesize that masked sub-series modeling can learn general-purpose representations.
These results indicate strong potential as a universal model for serving multiple domains.

\subsubsection{Fine-tuning with Varying Proportions of Training Set}
One of the most valuable characteristics of pre-trained models is their stronger generalization in tackling the label-sparse scenarios~\cite{brown2020language,cheng2023formertime}. We hence investigate the effectiveness of the TimeMAE model by simulating the data sparse scenarios with different proportions of training sets. It should be noted that all compared model variants preserve the same testing set to ensure a fair comparison. Accordingly, Figure~\ref{fig:ratio_training} shows the evaluation results of FineZero+ and the TimeMAE on datasets of PS and Epilepsy. As we can see, the performance substantially drops when a lower proportion of training data is used.  Notably, we find that our TimeMAE could achieve superior classification performance compared to randomly initialized models. Particularly in the dataset of Epilepsy, our TimeMAE could even achieve comparable classification performance tuned with very few labeled datasets, as the capacity of a randomly initialized version tuned with the full training set. This is meaningful in many scenarios of label sparsity.

\begin{figure*}[t]
	\centering
	\includegraphics[width=1.0\textwidth]{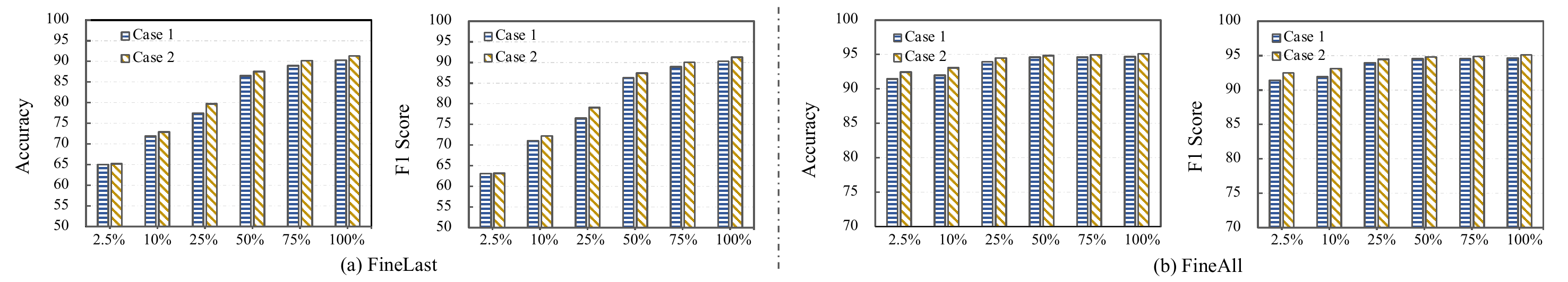} 
	\caption{Performance analysis of the TimeMAE model with respect to different proportions of the pre-training set (i.e., x-axis) while preserving the same fine-tuning set within case 1 (70\% of the full training set) or case 2 (full training set).}
	\label{fig:ratio_pre_training}
\end{figure*}
\subsubsection{Analysis of the Scalability of Model Size}
In the area of self-supervised pre-training, a series of works~\cite{devlin2018bert} demonstrate that stronger generalization models can be developed by jointly scaling the size of the pre-trained encoder or increasing the overall training duration. Inspired by these findings, we are interested in exploring whether similar performance can be achieved in representations of time series. Thus, we scale the model size of the encoder model by gradually varying the depth of encoder blocks ($l = 8, 16$), and the dimension of embedding size ($d = 64, 128$). Additionally, we control the number of epochs to verify whether longer pre-training can benefit the pre-training process ($n = 100, 200$). In our experiments, we establish six different models, equipped with various hyper-parameter settings. The specific experimental results are reported in Table~\ref{tab:scable_analysis}.  Overall, we observe that larger model sizes and longer pre-training epochs indeed can further generate superior representations with slightly improved generalization capacity on the HAR and Epilepsy datasets. In contrast, the bigger models obtain a significant performance gain on the PS datasets. This is probably because there are fewer pre-training sets on the PS dataset compared to the other two datasets. 
After all, a larger model should be matched well with a larger proportion of the training set so as to fulfill the expressive capacity of networks.

\subsubsection{Analysis of Varying Proportions of Pre-Training Set}
Next, we investigate whether the proposed model is susceptible to the saturation phenomenon reported in~\cite{kong2022understanding}, where enlarging the self-supervised pre-training set yields diminishing or even negligible performance gains. Given the limitations of available datasets, we assess this effect by varying the proportion of data used for pre-training while keeping the fine-tuning set unchanged. Concretely, we pre-train TimeMAE on two versions of the HAR dataset ($100\%$ and $70\%$), referred to as Case~1 and Case~2, and subsequently fine-tune both models on the same labeled subset. As illustrated in Figure~\ref{fig:ratio_pre_training}, increasing the pre-training set consistently enhances the downstream classification performance, with particularly clear improvements under linear evaluation. These observations highlight an encouraging property for practical deployment, as large volumes of unlabeled time series can be collected with minimal cost and without relying on expert annotation, allowing TimeMAE to benefit from broader unsupervised corpora. Consequently, effective foundation models can be built through learning from a vast scale of unlabeled time series datasets, thus achieving more promising results for serving target tasks.

\begin{figure}[h]
	\centering
	\includegraphics[width=0.5\textwidth]{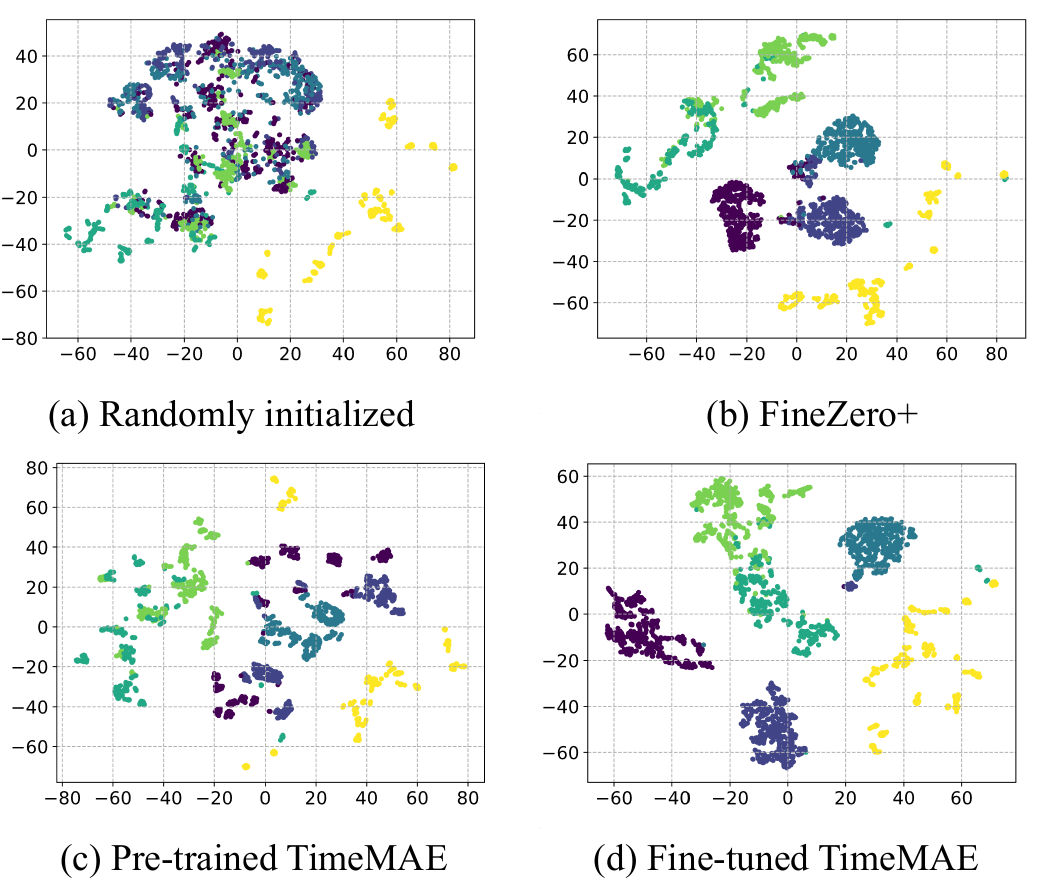} 
	\caption{T-SNE visualizations of feature vectors on the HAR dataset, in which each color denotes a specific class category.}
	\label{fig:tsne}
\end{figure}
\subsubsection{Visualization Analysis}
Next, we aim to understand the proposed approach through visualization analysis. To conserve space, we use T-SNE~\cite{van2008visualizing} over the HAR dataset to display the learned features along with their labels. Figure~\ref{fig:tsne}(a) provides the visualization of features derived from randomly initialized encoders, while Figure~\ref{fig:tsne}(b) describes the visualization of features following supervision training without pre-training enhancement. Figure~\ref{fig:tsne}(c) and (d) show the visualization results of extracted features from a frozen pre-trained encoder and a fine-tuned one in TimeMAE. We draw the following observations from the visualization results: it is possible to separate well the extracted features from the tuned TimeMAE and FineZero+. Among them, the Fine-tuned TimeMAE achieves better separation results. 
Such results demonstrate that TimeMAE enhances category separation and captures key features of continuous time series.
Visualized features extracted from the pre-trained TimeMAE can also be well separated. Such results suggest that the pre-trained representations could largely reflect the underlying features of raw time series data. 
These results show that TimeMAE guides category separation in the latent space and captures key features of continuous time series.

\section{Conclusion}
In this work, we proposed a novel self-supervised model, named TimeMAE, designed to learn representations of time series using transformer networks. The unique characteristics of the TimeMAE is its adoption of sub-series as basic semantic units, combined with a window slicing operation and masking strategies.  Several novel challenges associated with this innovative modeling paradigm were well solved. With TimeMAE, transferable time series representations can be effectively learned by leveraging more enriched basic semantic units and the bidirectional encoding strategy.  Extensive experiments were conducted on five public datasets. The results of these experiments illustrate the effectiveness of the proposed TimeMAE by reporting some insightful findings.
\section*{Acknowledgments}
This research was supported by grants from the National Natural
Science Foundation of China (No. 62502486, 62337001), the grants of the Provincial
Natural Science Foundation of Anhui Province (No. 2408085QF193), USTC Research Funds of the DoubleFirst-Class Initiative (No. YD2150002501),
the Fundamental Research Funds for the Central Universities of
China (No. WK2150110032).

\section*{Ethical Considerations}
All experiments in this study are conducted on publicly available, fully anonymized datasets that contain no personal or sensitive information. The research does not involve human subjects, protected attributes, or any data that could lead to privacy violations. No foreseeable negative societal impacts are identified, such as unfair treatment, malicious surveillance, or potential misuse by adversarial actors. The proposed method is designed for general-purpose time series analysis and does not target sensitive application domains. The work adheres to standard ethical guidelines for data usage, reproducibility, and responsible scientific research.


\bibliographystyle{ACM-Reference-Format}
\bibliography{main}

\end{document}